\documentclass{amia}
\usepackage{caption}
\usepackage{amsmath,amssymb}
\usepackage{booktabs}
\usepackage{tabularx}
\usepackage{float}
\usepackage{multirow}
\usepackage{subfigure}
\usepackage{subcaption}
\begin{document}

\title{Parameter-Efficient Token Embedding Editing for Clinical Class-Level Unlearning}

\author{Iyad Ait Hou$^1$, Shrenik Borad$^1$, Harsh Sharma$^1$, Pooja Srinivasan$^1$, Rebecca Hwa$^1$, Aya Zirikly$^1,^2$}

\institutes{
    $^1$Department of Computer Science, The George Washington University, Washington, DC
    $^2$Center for Speech and Language Processing, Johns Hopkins University, Baltimore, MD 
}

\maketitle

\section*{Abstract}

\textit{Machine unlearning is increasingly important for clinical language models, where privacy regulations and institutional policies may require removing sensitive information from deployed systems without retraining from scratch. In practice, deletion requests must balance effective forgetting of targeted information with preservation of model utility and minimal parameter modification.  We introduce Sparse Token Embedding Unlearning (STEU), a parameter-efficient method for behavioral class-level unlearning that updates only PMI-selected token embeddings together with a small classifier head while keeping all encoder layers frozen.  Across experiments on MIMIC-IV, MIMIC-III, and eICU using BioClinicalBERT, BERT-base, and DistilBERT, STEU consistently suppresses the target class while largely preserving retained task performance. In the primary MIMIC-IV setting, STEU achieves near-complete forgetting (forget F1 = 0.0004) while maintaining competitive retained utility (retain avg F1 = 0.4766) after modifying only 0.19\% of model parameters. These results suggest that targeted behavioral unlearning can be achieved through sparse embedding edits without modifying deeper encoder representations.}

\section*{Introduction}

Large language model (LLM)–based clinical text classifiers are increasingly used to analyze electronic health records (EHRs) and predict outcomes such as diagnostic categories, patient cohorts, or clinical risk indicators from medical notes. Because these systems are trained on sensitive clinical data, healthcare institutions may later require that specific information be removed from a trained model. For example, hospitals may request that models stop predicting a particular disease category, remove the influence of specific patient cohorts, or forget information originating from a particular dataset.

This problem is studied under the framework of \emph{machine unlearning}, which aims to remove the influence of selected training data from a trained model while preserving performance on the remaining data. Ideally, the resulting model should behave as if the removed data had never been observed during training. The obvious solution is to retrain the model from scratch without the target data, but this approach is often computationally expensive or infeasible when training data are distributed across institutions or cannot be re-accessed \cite{bourtoule2021sisa, guo2020cdr,wang2024survey}. Hence, researchers focus on algorithms that edit components of the classifiers to achieve unleanring. 

In a transformer-based classifier, the two potential components for unlearning are the embedding layer and the encoding layer: the embedding layer maps tokens to vector representations; the encoder layers transform these representations through multiple attention and feed-forward blocks to produce contextual features used for prediction. 
Existing unlearning approaches usually intervene at the encoder level, modifying a large portion of model parameters. While effective, such updates can be difficult to audit and may violate parameter-change constraints in shared deployments.

In contrast, the embedding layer typically contains only a small fraction of total model parameters; however, it still contains important information. Thus, our question is: can small, targeted edits at the embedding layer induce meaningful behavioral changes in the model? If specific clinical concepts are associated with a limited set of lexical signals, modifying the representations of those signals may be sufficient to alter downstream predictions.
To explore this idea, we introduce \textbf{Sparse Token Embedding Unlearning (STEU)}, a parameter-efficient method that performs class-level behavioral unlearning by editing a small subset of token embeddings selected using pointwise mutual information (PMI). These embedding edits are combined with a lightweight classifier head while keeping all encoder layers frozen.
\newpage
\textbf{Contributions}
This paper makes the following contributions:

\begin{itemize}
\item \textbf{Embedding-layer unlearning framework.}
We propose Sparse Token Embedding Unlearning (STEU), a method that performs behavioral class-level unlearning through targeted edits to a small subset of token embeddings rather than encoder-wide updates.

\item \textbf{Parameter-efficient unlearning mechanism.}
STEU modifies only a tiny fraction of model parameters, enabling localized and auditable edits suitable for deployment-constrained environments.

\item \textbf{Systematic empirical evaluation.}
We evaluate STEU across multiple transformer backbones and clinical datasets, demonstrating that localized embedding edits can achieve strong forgetting behavior while preserving retained task performance.
\end{itemize}

\begin{figure}
    \centering
    \includegraphics[width=1\linewidth]{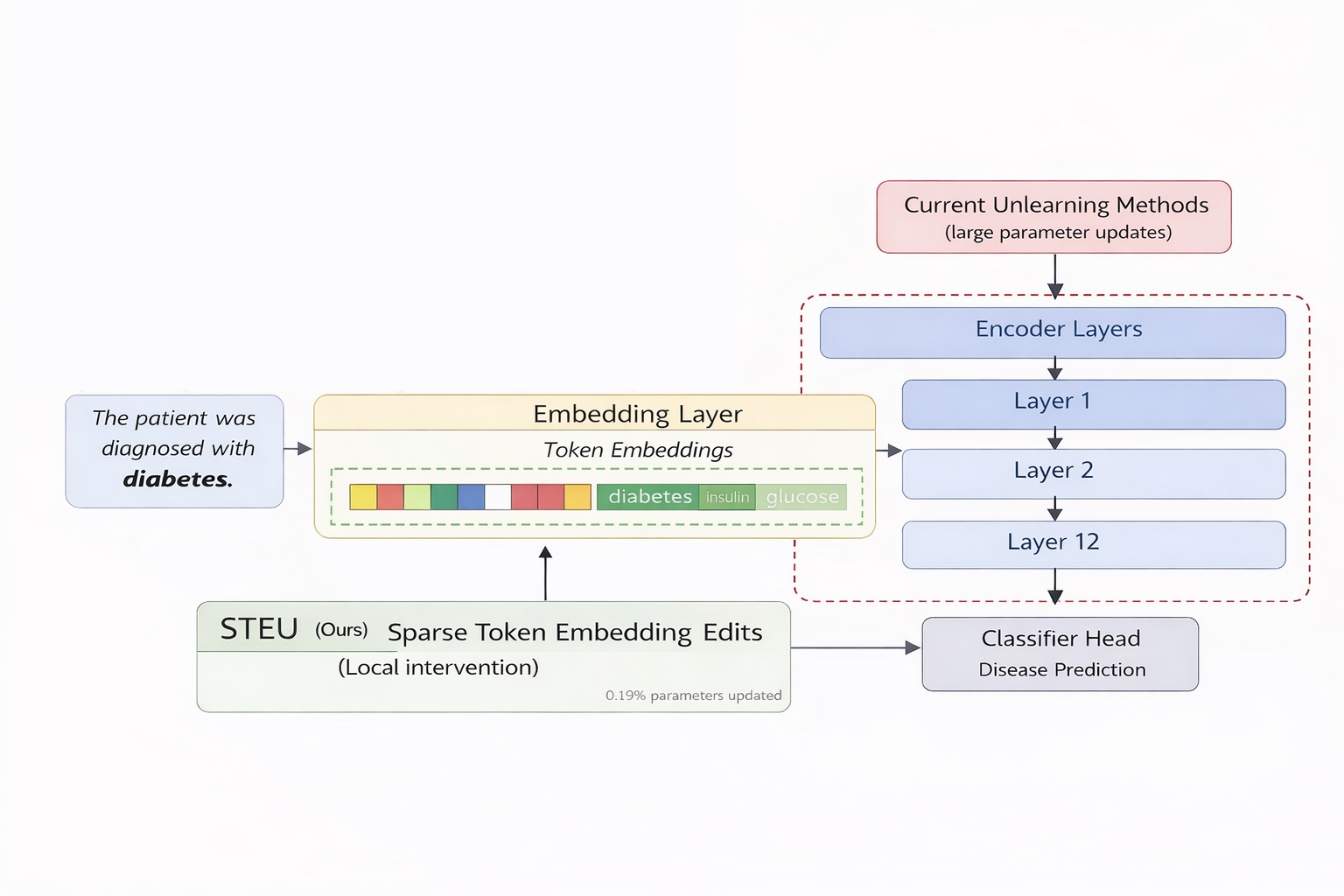}
    \caption{Conceptual overview of Sparse Token Embedding Unlearning (STEU). While existing machine unlearning methods modify large portions of encoder parameters, STEU performs localized edits to selected token embeddings, enabling class-level behavioral unlearning with minimal parameter updates.}
    \label{fig:placeholder}
\end{figure}

\section*{Related Work}

Machine unlearning methods span a spectrum from exact retraining to approximate post-hoc parameter updates. The most direct solution to a deletion request is retraining the model from scratch without the target data. However, this approach is often computationally prohibitive in modern deep learning systems and impractical in settings where training data are distributed or no longer accessible. Early practical approaches therefore explored partition-based retraining strategies such as SISA (Sharded, Isolated, Sliced, and Aggregated training) \cite{bourtoule2021sisa}, which reduce the cost of retraining by limiting the portion of the model that must be recomputed when deletion requests occur.

Another line of work focuses on providing formal guarantees for data removal. Certified data removal methods aim to ensure that the influence of a deleted sample on the trained model can be bounded or eliminated \cite{guo2020cdr}. These approaches provide theoretical guarantees but often rely on simplified model assumptions or operate primarily at the decision-boundary level in deep architectures. As a result, practical implementations frequently modify classifier heads or decision boundaries rather than deeply intervening in the representation layers of the model.

Influence-based approaches attempt to estimate the contribution of individual training samples to model parameters. Techniques based on influence functions approximate the effect of removing a training point using inverse-Hessian approximations \cite{koh2017influence}. While theoretically appealing, these methods can become unstable and computationally expensive in large transformer architectures due to the high dimensionality of model parameters.

More recently, machine unlearning has also been studied in distributed and federated settings, where deletion requests must be handled under additional communication and governance constraints. Federated unlearning methods such as FedEraser and related approaches extend the unlearning problem to decentralized training environments where data are partitioned across institutions \cite{wu2022federaser,romandini2024fu,jeong2024fedsok}.

Across these approaches, two widely used deep unlearning objectives have emerged. The first performs gradient ascent on the forget set in order to maximize loss on the data that should be removed. The second directly suppresses the target signal by pushing predictions for the forget set toward a neutral or zero output \cite{bourtoule2021sisa,wang2024survey}. In practice, these objectives are typically applied to large portions of the model, including encoder layers and the classifier head. While such interventions can achieve strong forgetting, they often require modifying a substantial fraction of model parameters and may introduce collateral degradation in retained task performance.

An emerging perspective is that the effectiveness and cost of unlearning depend critically on the \emph{intervention locus} within the model architecture. Head-only interventions are highly localized but limited in their ability to reshape internal representations. Encoder-level updates provide greater expressive power but involve broad parameter modifications that may be difficult to audit or deploy under parameter-change constraints.

This locus-aware perspective is further motivated by research on memorization and privacy leakage in deep models. Studies on membership inference and training data extraction demonstrate that neural models can encode sensitive information in distributed representations \cite{shokri2017membership,carlini2019secret,carlini2021extracting}. These findings highlight the importance of understanding where sensitive signals enter and propagate through the model.

In this work, we explore a different intervention locus for machine unlearning. Specifically, we target the \emph{embedding layer}, where lexical signals first enter the model representation. By editing a sparse subset of token embeddings while keeping the encoder frozen, our method introduces an intermediate design point between head-only and encoder-level updates. This embedding-level intervention enables localized behavioral unlearning with substantially smaller parameter updates than encoder-wide methods.

\section*{Methodology}

We study \emph{behavioral class-level unlearning} in clinical text classification. Our setting begins with a transformer-based classifier trained on electronic health record (EHR) notes to predict disease categories in a multi-label setting. After training, a deletion request specifies one target disease class that should be forgotten. For example, a model may predict diagnoses such as myocardial infarction, pneumonia, and diabetes from clinical notes. If a hospital later determines that the myocardial infarction labels were derived from an unreliable coding pipeline or inconsistently annotated, practitioners may wish to remove the model’s ability to predict that class without retraining the entire system. In this case, the model should stop assigning high probability to myocardial infarction—even when notes contain phrases such as ``acute myocardial infarction'' or ``cardiac catheterization''—while predictions for other conditions remain unaffected. The objective is therefore to suppress the target-class prediction behavior while preserving performance on all remaining classes and modifying as little of the model as possible.

 In our case, the deletion target is a disease category rather than an individual patient record. Accordingly, we adopt a practical behavioral definition of success: an unlearned model should achieve near-zero predictive behavior on the forget class while maintaining retained utility on non-target classes under a strict parameter budget. This is distinct from certified sample-level removal, which aims to guarantee that specific training examples have no remaining influence on the model.

Our method, \textbf{Sparse Token Embedding Unlearning (STEU)}, is motivated by the observation that many clinical concepts are associated with a relatively small set of lexical signals. Instead of modifying large portions of the encoder, STEU identifies tokens that are strongly associated with the forget class and restricts learning updates to their embedding vectors together with a lightweight classifier head (i.e., the final linear layer that maps encoder representations to disease label predictions). All encoder layers remain frozen. This produces a localized intervention that aims to maximize forgetting while minimizing collateral changes to the model.

\textbf{Problem Setting} Let $f_\theta$ denote a transformer-based multi-label classifier with parameters $\theta$, producing logits $z_\theta(x) \in \mathbb{R}^C$ for input text $x$ over $C$ disease classes. In our experiments, $f_\theta$ is instantiated with Bio\_ClinicalBERT, BERT-base, or DistilBERT \cite{alsentzer2019clinicalbert,devlin2019bert,sanh2019distilbert}. Each input document is a clinical note, and the task is to predict one or more disease categories from the note.

Given a forget class $c_f$, the goal is to obtain updated parameters $\theta'$ such that $f_{\theta'}$ no longer predicts $c_f$ on forget-class inputs while maintaining performance on all other classes. Let $\theta_0$ denote the baseline model parameters before unlearning. Our target is therefore not to erase all traces of training data in a certified sense, but to suppress target-class predictive behavior under fixed evaluation while preserving non-target utility and keeping the update surface small.

\textbf{Overview of STEU} STEU operates in two stages. First, it identifies a small set of tokens that are strongly associated with the forget class. Second, it restricts optimization to the embedding rows corresponding to those tokens together with the classifier head. The key idea is that if disease-specific behavior is partly driven by a sparse set of lexical cues, then targeted edits at the embedding layer may be sufficient to induce forgetting without modifying the encoder.

This design differs from broader unlearning approaches that update encoder blocks or large portions of the model. In STEU, the encoder remains completely fixed, and only a constrained set of parameters is allowed to change. This makes the intervention highly localized and directly measurable in terms of parameter budget.

\textbf{Token Selection via PMI} The first component of STEU is a token selector that identifies tokens carrying class-discriminative signal for the forget class $c_f$. We tokenize all forget-class and retain-class documents, compute token frequencies $n_f(t)$ and $n_r(t)$, and assign each token $t$ a frequency-weighted pointwise mutual information score:
\begin{equation}
\text{score}(t) =
\underbrace{\log_2\!\left(\frac{P(t \mid \text{forget})}{P(t \mid \text{all})}\right)}_{\text{PMI: class specificity}}
\times
\underbrace{\log(1+n_f(t))}_{\text{frequency weight}}.
\end{equation}

The PMI term measures how strongly a token is associated with the forget class relative to the full corpus, while the frequency term downweights rare tokens that appear too infrequently to drive class-level behavior at scale. Tokens below a minimum frequency threshold are excluded. The top-$k$ tokens ranked by this score form the selected set $\mathcal{S}$.

PMI is well suited to this setting for three reasons. First, it is label-aware and directly measures token-class association. Second, it is deterministic and therefore produces a reproducible token set across runs. Third, it naturally separates class-specific terminology from common shared vocabulary. For example, generic clinical terms such as ``patient'' or ``discharge'' receive low scores, while disease-specific terms associated with the forget class receive high scores.

\begin{figure}[H]
  \centering
  \includegraphics[width=0.85\textwidth]{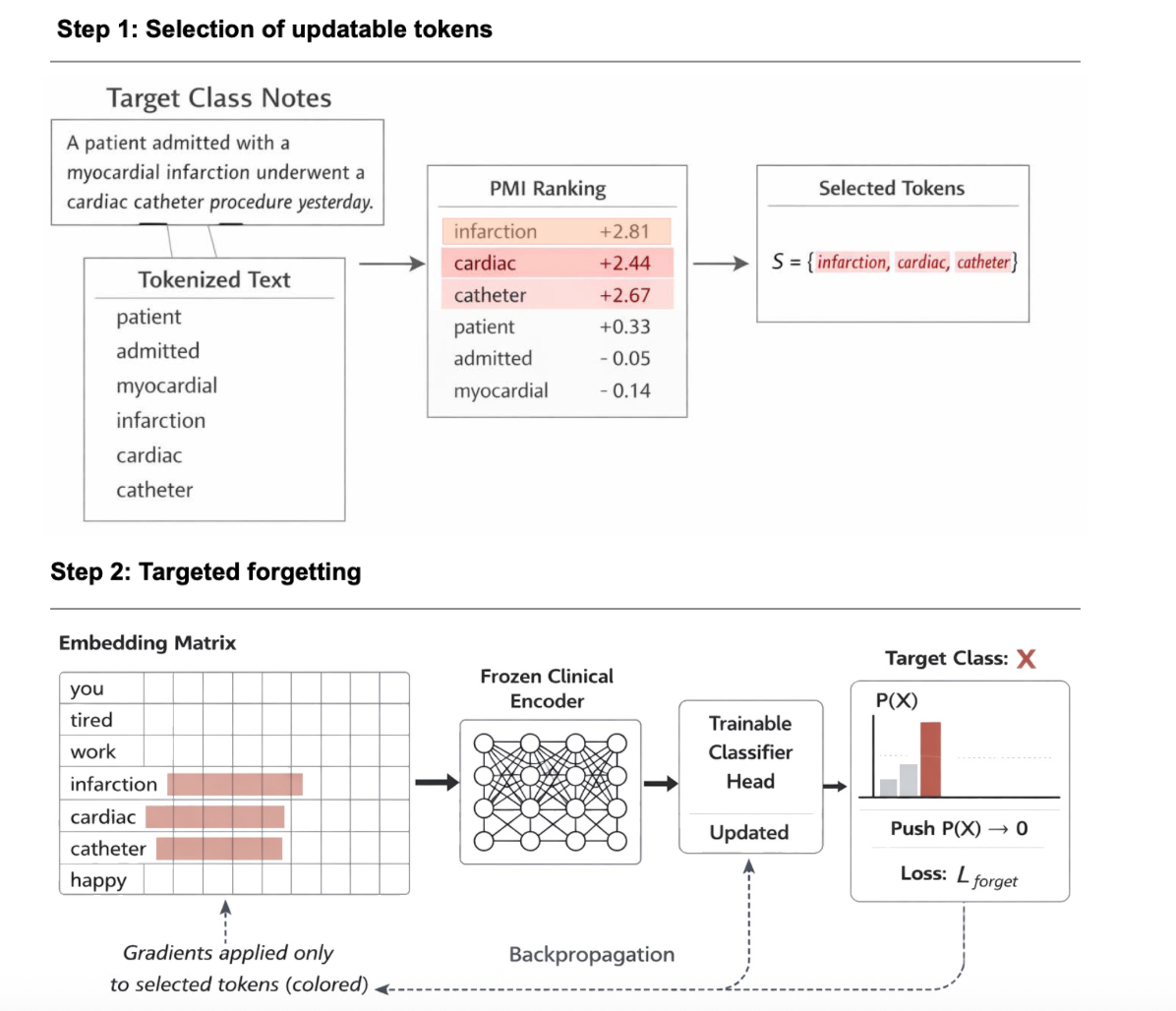}
  \caption{STEU workflow figure combining token selection and constrained unlearning. The pipeline first identifies class-discriminative tokens via frequency-weighted PMI, then updates only the selected embedding rows and classifier head with gradient masking while all encoder layers remain frozen.}
\end{figure}

\textbf{Constrained Update Surface} Once the token set $\mathcal{S}$ has been selected, STEU restricts all gradient updates to a small and explicitly defined subset of parameters. All encoder layers are frozen, including all attention and feed-forward blocks. Position embeddings and segment embeddings are also frozen. Only two parameter groups remain trainable:

\begin{enumerate}
    \item \textbf{Selected token embedding rows.} Let $\mathbf{E} \in \mathbb{R}^{V \times d}$ denote the word embedding matrix, where $V$ is the vocabulary size and $d$ is the embedding dimension. Only the rows of $\mathbf{E}$ corresponding to token IDs in $\mathcal{S}$ are allowed to update.
    
    \item \textbf{Classifier head.} We also allow the final linear classification layer $\mathbf{W}_c \in \mathbb{R}^{d \times C}$ and its bias to update. This provides a small amount of additional capacity to remap the perturbed representation into a suppressed output for the forget class.
\end{enumerate}

During backpropagation, we apply a gradient mask to the embedding matrix so that all rows $i \notin \mathcal{S}$ receive zero gradient. This ensures that only the selected token embeddings are modified. The result is a sharply constrained update surface consisting of selected lexical representations plus the lightweight classification head.

\textbf{Training Objective} Each training step uses one forget batch $\mathcal{B}_f$ and one retain batch $\mathcal{B}_r$. The full objective combines a forget loss and a retain-utility loss:
\begin{equation}
\mathcal{L}
=
\lambda_u \cdot \mathcal{L}_{\text{forget}}
+
\lambda_k \cdot \mathcal{L}_{\text{utility}}.
\end{equation}

For the forget loss, we use direct suppression of the target class by forcing its target label to zero:
\begin{equation}
\mathcal{L}_{\text{forget}}
=
\text{BCE}\!\left(z_{\theta'}(\mathcal{B}_f),\; y^{(0)}\right),
\end{equation}
where $y^{(0)}$ denotes a modified target vector in which the forget-class label is set to zero. This objective explicitly trains the model to stop producing positive predictions for $c_f$ on forget-class inputs.

To preserve non-target behavior, we anchor the updated model to the frozen baseline on retain data:
\begin{equation}
\mathcal{L}_{\text{utility}}
=
\frac{1}{C-1}
\sum_{c \neq c_f}
\text{BCE}\!\left(
\sigma(z_{\theta',c}(\mathcal{B}_r)),
\;
\sigma(z_{\theta_0,c}(\mathcal{B}_r))
\right).
\end{equation}

This retain objective encourages the updated model to preserve the baseline decision profile for all non-target classes. In other words, the model is pushed to forget the target class while remaining behaviorally close to the original classifier on the rest of the task.

After computing the loss, we backpropagate gradients as usual. Before the optimizer step, the embedding gradient mask is applied so that only rows in $\mathcal{S}$ remain trainable.

\textbf{Why Embeddings Plus Head} A natural question is why STEU updates both selected embeddings and the classifier head rather than modifying only one of these components. Updating the classifier head alone provides limited control over the lexical signals that trigger the forget class. The encoder still produces strong hidden representations for tokens associated with the concept, and the linear head must attempt to suppress these signals globally. This often interferes with nearby decision boundaries and can degrade predictions for related classes.

Conversely, embedding-only updates can disrupt the lexical signal entering the model, but the frozen classifier head still maps the resulting hidden states using its original parameters. In practice, this limits the model's ability to fully suppress the forget-class output.

STEU therefore combines both interventions. Selected embeddings weaken the representation of the forget concept at the input level, while the classifier head adapts the final mapping from representation to label space. This introduces only a small number of additional parameters but substantially improves forgetting performance. In our experiments, embedding-only updates leave substantial residual forget-class behavior, whereas adding the head reduces forget F1 from $0.276$ to $0.0004$ with only a very small increase in parameter count.

\section*{Experimental Setup}

\textbf{Task Definition.}
We study \emph{behavioral class-level unlearning} in clinical text classification. Given a trained multi-label classifier and a deletion request specifying one disease category, the objective is to suppress predictions for the target class while preserving predictive performance on all remaining classes and modifying as few model parameters as possible. In each experiment, one disease category is designated as the \emph{forget class} and the remaining categories form the \emph{retain set}. We measure success by (i)~forget-class F1 driven to near zero and (ii)~bounded degradation in retain average F1 across non-target classes, under an explicit parameter-edit budget.

\textbf{Datasets and Label Space.}
We evaluate on three widely used electronic health record datasets: MIMIC-IV~\cite{johnson2023mimiciv}, MIMIC-III~\cite{johnson2016mimiciii}, and eICU~\cite{pollard2018eicu}. Diagnosis labels are mapped to categories defined by the \emph{Clinical Classifications Software Refined (CCSR)} system~\cite{ahrq_ccsr}, which aggregates thousands of fine-grained ICD-10-CM codes into clinically meaningful disease groups. From the full CCSR taxonomy we select six high-prevalence target groups (CCSR\_401, CCSR\_272, CCSR\_250, CCSR\_E87, CCSR\_V58, CCSR\_E78) to define a controlled multi-label classification task with realistic class imbalance. For each dataset, we subsample 50{,}000 training and 20{,}000 validation examples using patient-disjoint splits. Table~\ref{tab:datasets} summarizes the data sources.

\begin{table}[H]
\centering
\caption{Dataset summary. Experiments use 50{,}000/20{,}000 train/val subsamples with patient-disjoint splits from each corpus.}
\label{tab:datasets}
\small
\begin{tabularx}{\linewidth}{lXXX}
\toprule
\textbf{Dataset} & \textbf{Source} & \textbf{Total Admissions} & \textbf{Target Classes} \\
\midrule

MIMIC-IV & Discharge summaries \cite{johnson2023mimiciv,johnson2023mimicivnote} & 148{,}399 & Selected 6 CCSR groups \\
MIMIC-III & Discharge summaries \cite{johnson2016mimiciii} & 52{,}726 & Selected 6 CCSR groups \\
eICU & ICU clinical notes \cite{pollard2018eicu} & 200{,}859 & Selected 6 CCSR groups \\
\bottomrule
\end{tabularx}
\end{table}

\textbf{Pre-trained Encoders.}
We train baseline classifiers on each dataset using three pre-trained transformer encoders of varying capacity (Table~\ref{tab:backbones}). During unlearning, all encoder layers remain frozen; only selected token embedding rows and the classifier head are updated.

\begin{table}[H]
\centering
\caption{Pre-trained encoder configurations. Val Macro F1 is reported on the MIMIC-IV validation split.}
\label{tab:backbones}
\small
\begin{tabularx}{\linewidth}{lXXXX}
\toprule
\textbf{Encoder} & \textbf{Total Params} & \textbf{Layers} & \textbf{Hidden Dim} & \textbf{Val Macro F1} \\
\midrule
Bio\_ClinicalBERT \cite{alsentzer2019clinicalbert} & 108.3M & 12 & 768 & 0.499 \\
BERT-base \cite{devlin2019bert} & 109.5M & 12 & 768 & 0.484 \\
DistilBERT \cite{sanh2019distilbert} & 66.4M & 6 & 768 & 0.242 \\
\bottomrule
\end{tabularx}
\end{table}

\textbf{Experimental Protocol.}
For each encoder--dataset combination, we (1)~train a baseline multi-label classifier on the 50{,}000-sample training split, (2)~designate one CCSR category as the forget class, (3)~apply the unlearning method for 5 epochs, and (4)~evaluate forget F1 and retain average F1 on the held-out validation split.

\textbf{Comparison Methods.}
We compare STEU against three established unlearning approaches. 
\textit{Gradient ascent} maximizes the loss on the forget set by updating encoder layers 10--12 together with the classifier head (21.3M parameters, 19.6\% of the model)~\cite{bourtoule2021sisa}. 
\textit{Direct suppression} instead minimizes binary cross-entropy against zero labels for the forget class while using the same update surface (encoder layers 10--12 and classifier head)~\cite{wang2024survey}. 
\textit{Influence-weighted unlearning} applies approximate influence scores to weight gradient updates on the forget set across the same encoder layers (21.3M parameters)~\cite{koh2017influence}. 
We exclude head-only certified removal~\cite{guo2020cdr}, as it targets a different objective—decision-boundary adjustment rather than behavioral class-level unlearning—and intervenes at a different locus.

\section*{Results}

We evaluate STEU along three dimensions: (1) effectiveness of target-class forgetting, (2) preservation of predictive utility on non-target classes, and (3) parameter efficiency relative to encoder-level unlearning baselines. Forgetting performance is measured using the F1 score of the target class (``forget F1''), while retained utility is measured as the macro-averaged F1 across all remaining classes (``retain F1''). All deltas are computed as $\Delta = \text{Final} - \text{Baseline}$. Unless otherwise specified, experiments use top-$k=256$ PMI-selected tokens with embedding and classifier-head updates for five epochs.

\textbf{Method comparison.} We first compare STEU against representative encoder-level unlearning baselines using BioClinicalBERT on MIMIC-IV. Table~\ref{tab:method_cmp} reports forgetting effectiveness, retained utility, and the number of parameters updated by each method.

\begin{table}[t]
\caption{Comparison of forgetting effectiveness, retained utility, and parameter budget.}
\centering
\small
\begin{tabular}{lcccc}
\toprule
Method & Forget F1 & Retain F1 & Params Updated & \% Model \\
\midrule
Gradient ascent & 0.0000 & 0.3127 & 21.3M & 19.6\% \\
Direct suppression & 0.0000 & 0.4912 & 21.3M & 19.6\% \\
Influence-weighted & 0.0000 & 0.3228 & 21.3M & 19.6\% \\
\textbf{STEU (ours)} & \textbf{0.0004} & \textbf{0.4766} & \textbf{201K} & \textbf{0.19\%} \\
\bottomrule
\end{tabular}
\label{tab:method_cmp}
\end{table}

Encoder-level methods achieve complete forgetting but require modifying large portions of the model, updating over 21M parameters. STEU achieves near-complete forgetting while modifying more than two orders of magnitude fewer parameters. Despite the drastically smaller intervention, retained performance remains competitive.

\textbf{Cross-encoder / cross-dataset evaluation.} To evaluate robustness, we apply STEU across three clinical datasets (MIMIC-IV, MIMIC-III, and eICU) and three encoder architectures (BioClinicalBERT, BERT-base, and DistilBERT). Table~\ref{tab:steu_matrix} reports baseline and post-unlearning performance for all nine encoder–dataset combinations.

\begin{table}[t]
\caption{Cross-dataset and cross-encoder evaluation of STEU. The final column reports retained utility loss ($\Delta = \text{Final} - \text{Baseline}$).}
\centering
\small
\begin{tabular}{llccccc}
\toprule
Dataset & Encoder &
Forget Base & Forget Final &
Retain Base & Retain Final &
$\Delta$ Utility Loss \\
\midrule
MIMIC-IV & BioClinicalBERT & 0.4685 & 0.0000 & 0.5064 & 0.5038 & -0.0026 \\
MIMIC-IV & BERT-base & 0.4188 & 0.0000 & 0.4975 & 0.4921 & -0.0054 \\
MIMIC-IV & DistilBERT & 0.3521 & 0.0000 & 0.4413 & 0.4350 & -0.0063 \\

MIMIC-III & BioClinicalBERT & 0.4210 & 0.0000 & 0.6296 & 0.6261 & -0.0035 \\
MIMIC-III & BERT-base & 0.0393 & 0.0000 & 0.2750 & 0.2698 & -0.0052 \\
MIMIC-III & DistilBERT & 0.0285 & 0.0000 & 0.2147 & 0.2091 & -0.0056 \\

eICU & BioClinicalBERT & 0.9669 & 0.0043 & 0.9434 & 0.9417 & -0.0017 \\
eICU & BERT-base & 0.9125 & 0.0000 & 0.0966 & 0.0941 & -0.0025 \\
eICU & DistilBERT & 0.8732 & 0.0000 & 0.0814 & 0.0789 & -0.0025 \\
\bottomrule
\end{tabular}

\label{tab:steu_matrix}
\end{table}

Across all conditions, STEU suppresses the forget-class signal while largely preserving predictive performance on remaining classes. Retain deltas remain small, typically within a few thousandths of baseline performance.

\begin{figure}[H]
\centering
\includegraphics[width=0.65\linewidth]{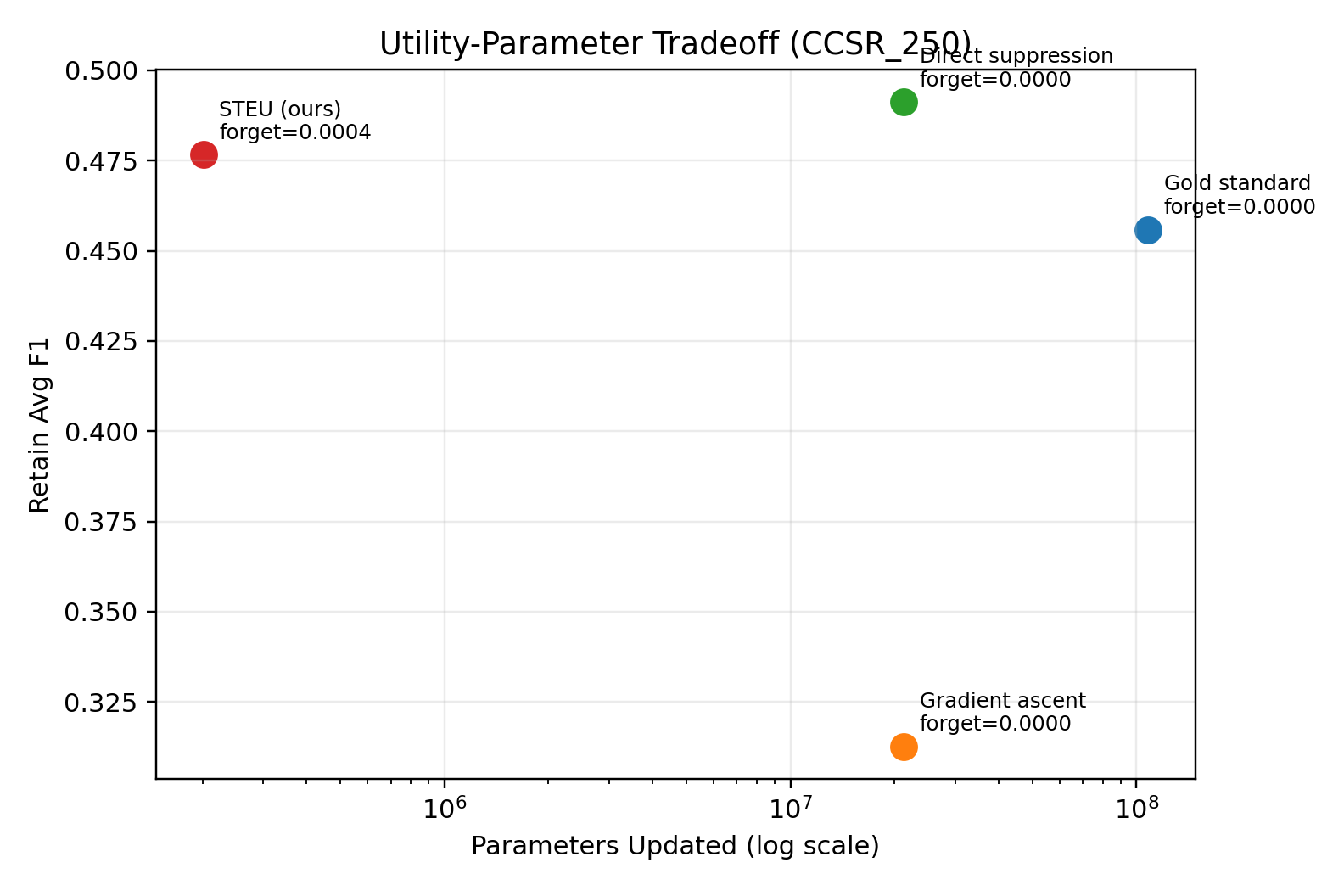}
\caption{Utility–parameter tradeoff. Each point shows retained F1 versus number of updated parameters (log scale), with forget F1 annotated. STEU occupies the high-utility, low-parameter region.}
\label{fig:tradeoff}
\end{figure}

\textbf{Utility–parameter tradeoff.} Figure~\ref{fig:tradeoff} visualizes the relationship between retained utility and the number of updated parameters. STEU occupies the high-utility, low-parameter region of the space, demonstrating that targeted embedding updates can achieve strong behavioral unlearning without modifying deeper encoder representations.

\textbf{Token-budget ablation.} Finally, we examine how the PMI token budget and classifier-head updates affect unlearning performance. Table~\ref{tab:ablation} reports results for several token budgets.

\begin{table}[t]
\caption{STEU token budget ablation.}
\centering
\small
\begin{tabular}{lcccc}
\toprule
Configuration & Params Updated & \% Model & Forget F1 & Retain Avg F1 \\
\midrule
$k=256$ (emb only) & 196,608 & 0.18\% & 0.2763 & 0.4398 \\
$k=512$ (emb only) & 393,216 & 0.36\% & 0.2771 & 0.4405 \\
$k=1024$ (emb only) & 786,432 & 0.71\% & 0.2766 & 0.4394 \\
\textbf{$k=256$ + head} & \textbf{201,222} & \textbf{0.19\%} & \textbf{0.0004} & \textbf{0.4766} \\
\bottomrule
\end{tabular}

\label{tab:ablation}
\end{table}

Increasing the token budget beyond $k=256$ provides little additional benefit, indicating that the highest PMI-ranked tokens capture most of the class-discriminative signal. Updating the classifier head is the critical component enabling complete suppression of the forget-class predictions.

\textbf{Limitations} This study has several limitations. First, our objective is \textit{behavioral class-level unlearning}: reducing target-class predictive behavior under a fixed evaluation setting. This does not constitute certified sample-level removal or a formal guarantee of zero influence from specific training examples.

Second, influence- and Fisher-style removal analyses are computationally expensive for transformer checkpoints and sensitive to Hessian approximations. Our influence-weighted baseline should therefore be interpreted as a comparative approximation rather than a certified removal method.

Third, experiments currently report single-seed runs and a limited set of baselines relative to the broader machine unlearning literature. In particular, partitioned retraining methods such as SISA are not fully explored in this work.

Fourth, PMI is used as a simple, deterministic, and interpretable token-selection heuristic. Future work should compare PMI with gradient-based attribution and Fisher-style token importance to evaluate selector sensitivity.

Finally, external dataset checks currently rely on rapid subset pipelines (e.g., ICD-prefix proxy groupings). Full label-harmonized replication across clinical datasets remains future work. The proposed method is designed for classification tasks and does not directly extend to generative settings where embedding perturbations may degrade fluency.

\section*{Conclusion} We introduce STEU, a lightweight approach to behavioral class-level unlearning in clinical text classification. The method identifies class-discriminative tokens using PMI and restricts gradient updates to those embedding rows together with a classifier head. Across multiple datasets and encoder architectures, STEU achieves near-complete forgetting while modifying fewer than 0.2\% of model parameters and largely preserving retained utility.

These results suggest that much of the class-specific predictive signal in transformer classifiers resides in a small subset of token embeddings rather than in deeper encoder representations. By targeting this subspace directly, STEU enables efficient behavioral suppression without destabilizing the broader model.

Future work should extend evaluation across additional datasets, incorporate multi-seed experiments and stronger unlearning baselines, and investigate privacy guarantees through attack- and influence-based analyses. Extending the approach to generative models and patient-level deletion protocols also remains an important direction.

\newpage
% References
\makeatletter
\renewcommand{\@biblabel}[1]{\hfill #1.}
\makeatother
\renewcommand{\refname}{\centering References}
\bibliographystyle{vancouver}
\bibliography{amia}

\end{document}